\documentclass{article}
\usepackage{ijcai25}
\usepackage{times} %
\usepackage{soul}
\usepackage{url}
\usepackage[hidelinks]{hyperref}
\usepackage[utf8]{inputenc}
\usepackage[small]{caption}
\usepackage{graphicx}
\usepackage{amsmath}
\usepackage{amsthm}
\usepackage{booktabs}
\usepackage{algorithm}
\usepackage{algorithmic}
\usepackage[switch]{lineno}

\usepackage{xspace}

\title{SplitQuantV2: Enhancing Low-Bit Quantization of LLMs Without GPUs}

\author{
Jaewoo Song$^{1}$
\And
Fangzhen Lin$^1$\\
\affiliations
$^1${Department of Computer Science and Engineering,\\The Hong Kong University of Science and Technology}\\
\emails
jsongab@connect.ust.hk,
flin@cse.ust.hk
}

\begin{document}
\newcommand{\splitquant}{SplitQuant\xspace}
\newcommand{\splitquantvii}{SplitQuantV2\xspace}
\maketitle
\begin{abstract}
The quantization of large language models (LLMs) is crucial for deploying them on devices with limited computational resources.
While advanced quantization algorithms offer improved performance compared to the basic linear quantization, they typically require high-end graphics processing units (GPUs), are often restricted to specific deep neural network (DNN) frameworks, and require calibration datasets.
This limitation poses challenges for using such algorithms on various neural processing units (NPUs) and edge AI devices, which have diverse model formats and frameworks.
In this paper, we show \splitquantvii, an innovative algorithm designed to enhance low-bit linear quantization of LLMs, can achieve results comparable to those of advanced algorithms.
\splitquantvii preprocesses models by splitting linear and convolution layers into functionally equivalent, quantization-friendly structures.
The algorithm's platform-agnostic, concise, and efficient nature allows for implementation without the need for GPUs.
Our evaluation on the Llama 3.2 1B Instruct model using the AI2's Reasoning Challenge (ARC) dataset demonstrates that \splitquantvii improves the accuracy of the INT4 quantization model by 11.76\%p, matching the performance of the original floating-point model.
Remarkably, \splitquantvii took only 2 minutes 6 seconds to preprocess the 1B model and perform linear INT4 quantization using only an Apple M4 CPU.
\splitquantvii provides a practical solution for low-bit quantization on LLMs, especially when complex, computation-intensive algorithms are inaccessible due to hardware limitations or framework incompatibilities.
\end{abstract}

\section{Introduction}
Large language models (LLMs) have seen rapid advancements in recent years.
Their impressive performance in natural language processing has sparked significant interest across diverse fields, including both research communities and industries.
However, LLMs are characterized by their large file sizes, high memory consumption, and substantial computing power requirements.
These challenges have led to an increasing focus on quantization techniques designed to reduce storage and memory demands while accelerating processing speeds.

Among many quantization techniques, linear quantization is the de facto standard method for quantization~\cite{gholami2022survey}.
It works by mapping original floating-point values to integers in a linear fashion.
Due to its simplicity and satisfactory performance on small to medium-sized deep neural network (DNN) models, linear quantization remains widely used in industry.
Nevertheless, research has demonstrated that linear quantization performs inadequately on LLMs, particularly in low-bit scenarios such as INT4 quantization~\cite{frantar2022gptq}.

To address the challenges associated with LLM quantization, more advanced methods have been developed~\cite{yao2022zeroquant}~\cite{frantar2022optimal}~\cite{frantar2022gptq}.
However, these techniques face three major issues that limit their practicality.

First, they demand high-performance GPUs due to their computational intensity.
Ironically, the very environments that lack powerful GPUs, which will thus benefit most from quantization, are unable to perform the quantization process by themselves because of their hardware limitations.

Second, these methods are often implemented exclusively on popular frameworks designed for LLM inference, such as Hugging Face Transformers~\cite{wolf-etal-2020-transformers}.
While these frameworks are user-friendly for inference, they pose difficulties for researchers and developers who wish to customize model structures extensively.
Additionally, companies that design their own neural processing units (NPUs) or DNN frameworks must implement quantization methods independently, which is challenging due to the complexity of these advanced techniques.

Third, the execution of these algorithms requires calibration datasets~\cite{liu2024spinquant}~\cite{frantar2022gptq}.
This requirement further restricts their applicability in situations where sufficient calibration data is unavailable.

Given these challenges, \splitquantvii presents a promising solution.
Built on the \splitquant method~\cite{song2025splitquant}, which was designed to enhance quantization for small language and computer vision models on edge devices, \splitquantvii extends these capabilities to LLMs.
It serves as a preprocessing technique that facilitates the quantization of LLMs, enabling effective results with basic linear quantization without the need for GPUs.

\splitquantvii enhances quantization resolution by dividing each linear layer (as well as convolution layers for computer vision models) into three mathematically equivalent layers.
The weights and biases of the original layer are partitioned using k-means clustering to improve quantization resolution.
Unlike its predecessor, \splitquantvii does not split activation layers, as quantizing these layers requires calibration datasets, which are often unavailable in the scenarios targeted by \splitquantvii.
Additionally, numerous practical improvements have been made to the code compared to \splitquant, ensuring that \splitquantvii is easily usable with PyTorch~\cite{paszke2019pytorch}.
When applied to the Llama 3.2 1B Instruct model~\cite{dubey2024llama}, \splitquantvii enhanced the accuracy of INT4 quantization model by 11.76\%p on the ARC dataset prepared for Llama 3.2 1B Instruct by Meta~\cite{llama32eval}, matching the accuracy of the original floating-point model.

\splitquantvii is lightweight, easy to implement, and highly portable LLM preprocessing method for enhancing quantization.
It is particularly beneficial for improving quantization results in environments lacking GPU access.
It can be easily adapted for specific NPUs or frameworks, and is effective even when calibration datasets are unavailable.
To our knowledge, \splitquantvii is the first approach to make linear quantization feasible for LLMs through model restructuring.
With its simplicity and robust performance, we believe \splitquantvii will inspire further research into preprocessing model structures, making linear quantization for LLMs more feasible, accessible, and accurate.

\section{Related Works}
\subsection{Linear Quantization}
Linear quantization is one of the most basic and straightforward quantization methods, often favored for its simplicity and ease of implementation~\cite{gholami2022survey}.
It works by mapping a range of continuous floating-point values in $[\beta, \alpha]$ to integer values in the range $[-2^{b-1}, 2^{b-1}-1]$ where $b$ is the target bit-width $b$.
For an original value $x$, the quantized value, denoted as $Q(x)$, is calculated as follows:%
\begin{eqnarray}
Q(x)&=&\mathrm{INT}\left( Sx \right) + Z\\
S&=&\frac{2^{b}-1}{\alpha - \beta}\\
Z&=&-2^{b-1} - \mathrm{INT}\left( S \beta \right)
\end{eqnarray}%
where $S$ is the scaling factor, $\mathrm{INT}()$ is a rounding function and $Z$ is the zero-point serving as an offset.

The scaling factor $S = \frac{2^{b}-1}{\alpha - \beta}$ maps original values in $[\beta, \alpha]$ to quantized values in $[-2^{b-1}, 2^{b-1}-1]$.
It is important to note that larger scaling factors result in better quantization resolution, while smaller scaling factors lead to poorer resolution.

Linear quantization is vulnerable to the influence of outliers, which are input values significantly deviating from the mean.
These outliers distort the quantization process by increasing the denominator of the scaling factor, $\alpha - \beta$, thus reducing the scaling factor and ultimately degrading the quantization resolution.

In linear quantization, weights and biases can be effectively quantized without a calibration dataset.
However, for activation quantization, a calibration dataset is necessary to estimate the minimum and maximum activation values.

Linear quantization is generally effective for small and medium-sized models where the impact of outliers is less pronounced.
However, its performance diminishes significantly when applied to LLMs.

\splitquantvii significantly enhances the accuracy of linear quantization for LLMs. As detailed in the following section, applying \splitquantvii to the linear INT4 quantization of the Llama 3.2 1B Instruct model on the ARC dataset yielded an 11.76\%p improvement and achieved the accuracy of the original floating-point model.

\subsection{LLM Quantization Algorithms}
The limitations of linear quantization in quantizing the LLMs have led to the development of more advanced quantization algorithms specifically designed for LLMs.
These advanced algorithms involve computation-intensive processes to enhance quantization outcomes.
For example, they might employ specific quantization granularities, conduct layer-wise knowledge distillation, and repeatedly calculate Hessian matrices.

Due to their computational demands, these algorithms heavily depend on GPUs, utilizing GPU-specific calculation kernels and executing numerous operations in batch mode.
Despite this, they can be time-consuming; for instance, ZeroQuant requires 3.1 hours on a single A100 GPU to quantize an LLM with 1.3 billion parameters.
This dependence on high-performance computing resources poses a challenge for environments lacking powerful GPUs.
Additionally, these algorithms are often integrated into popular LLM frameworks like Hugging Face, which can limit their adaptability to proprietary NPUs and other frameworks.
 
 In contrast, \splitquantvii offers a significantly faster alternative.
 Without the use of GPUs, \splitquantvii completed preprocessing in 1 minute and 56 seconds, with an additional 7 seconds for linear INT quantization of the Llama 3.2 1B Instruct model, using only an Apple M4 CPU.
 Compared to ZeroQuant's 3.1 hours on an A100 GPU for quantizing a 1.3B model, this represents a substantial improvement.
 
\subsection{Model Restructuring}
Since \splitquantvii is built upon the foundation of \splitquant ~\cite{song2025splitquant}, the related works concerning model restructuring are largely similar to those associated with \splitquant.

Net2Net was the pioneering work that introduced the concept of function-preserving transformations in DNNs~\cite{chen2015net2net}.
Its main goal was to improve DNN training by expanding the parameter space, rather than focusing on quantization.

The Cell Division and OCS methods adapt the Net2Net approach for quantization, but they have been applied exclusively to computer vision models~\cite{park2019cell}~\cite{zhao2019improving}.
Cell Division reduces the bit-width of weight parameters, while OCS mitigates the impact of outliers by duplicating neurons and halving their outgoing weights or outputs.

The main distinction between \splitquantvii and these earlier methods is that \splitquantvii employs k-means clustering to effectively partition layer weights.
Furthermore, \splitquantvii is specifically tailored for LLMs, in contrast to Cell Division and OCS, which target computer vision models with significantly fewer parameters.
Notably, \splitquantvii can improve quantization resolution even in the absence of outliers, whereas OCS primarily addresses outliers.

\section{Algorithm and Methodology}
\begin{figure*}[!tb]
    \centering
    \includegraphics[width=0.8\textwidth]{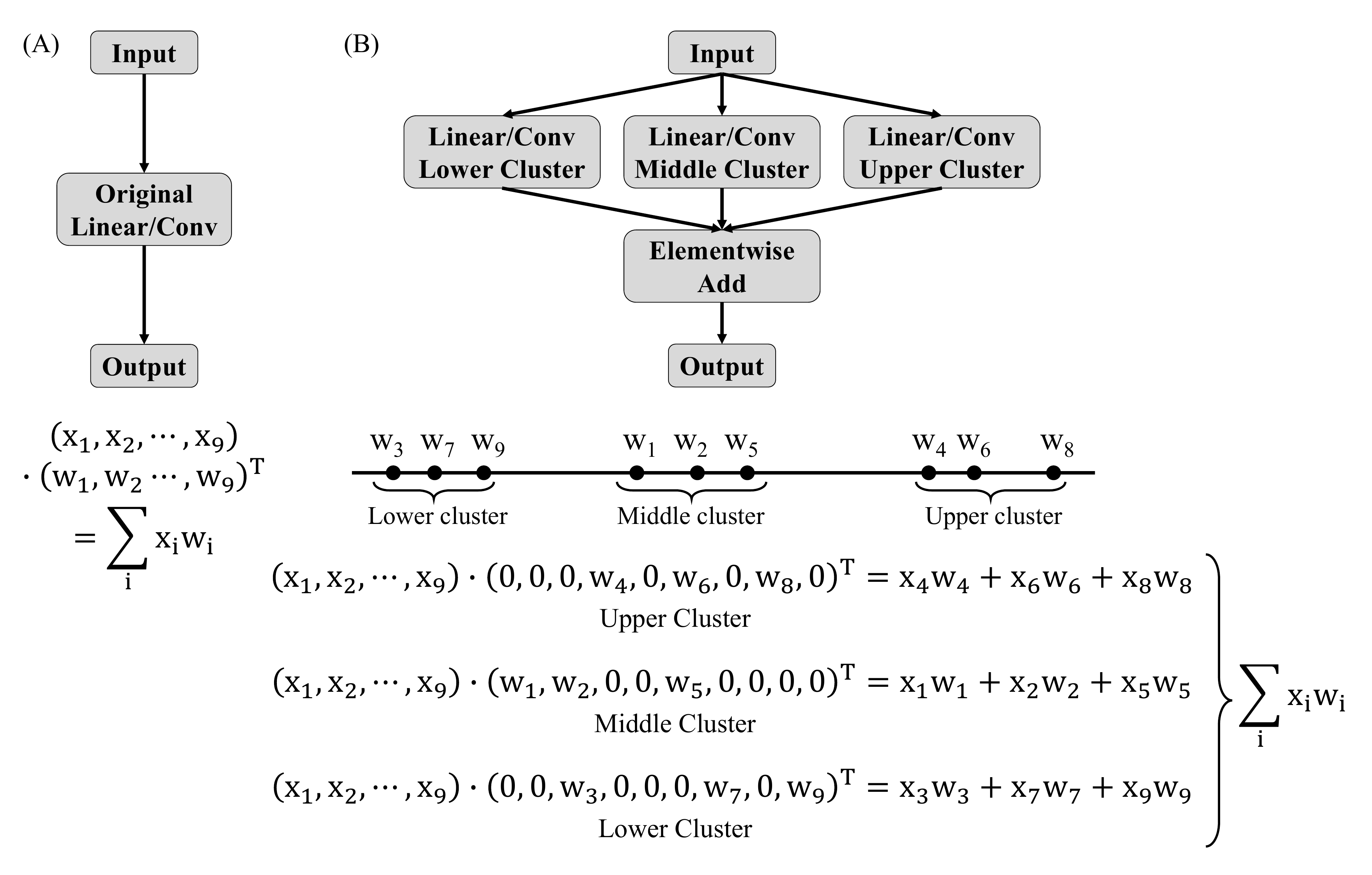}
    \caption{Modified and redrawn from \splitquant [Song and Lin, 2025]. (A) Original linear or convolution layer. (B) \splitquant improves quantization resolution by using k-means clustering on weights and biases to split the original layer into lower, middle and upper cluster layers. The functionality of the original layer is preserved.}
    \label{fig:splitquant}
\end{figure*}

The development of \splitquantvii builds upon the foundational concepts introduced by the \splitquant method.
For layers with weights, \splitquant pioneered the use of k-means clustering to split the layer, thereby achieving high quantization resolution while preserving the functionality of the original layer as shown in Figure~\ref{fig:splitquant}.
This technique clusters weights and biases into lower, middle, and upper groups, with outliers typically assigned to the lower and upper clusters.
Because the range of weight values in each split layer is much narrower than the original range, the scaling factor for each split layer increases and therefore quantization resolution is enhanced.

While \splitquant can be applied to any type of layer with weights, its primary focus is on linear and convolutional layers.
Since the Query, Key, and Value components within the attention heads of transformer blocks are implemented as linear layers, \splitquant extends its applicability to transformers as well.

\splitquantvii diverges from its predecessor by not splitting activation layers, concentrating solely on weight quantization.
This decision aligns with the algorithm's target applications where calibration datasets are unavailable.

In \splitquantvii, the k value for k-means clustering is fixed at 3.
This choice is based on the observation that increasing the number of clusters beyond three does not yield significant benefits.
Instead, it will increase the size of the quantized model excessively, negating the advantages of quantization.

Popular frameworks like PyTorch and HuggingFace represent embeddings of LLMs by weights.
However, embedding layers serve the functionality of lookup tables, which is different from linear or convolutional layers.
Therefore, \splitquantvii does not split embedding layers.

Similarly, PyTorch represents the gamma and beta parameters of batch normalization layers as weights and biases, with other normalization layers implemented in a similar manner.
\splitquantvii also does not split normalization layers.
It is worth noting that normalization layers can be easily folded into the preceding linear or convolution layers to simplify DNNs before applying \splitquantvii.

\section{Evaluation}
The performance of \splitquantvii was assessed using the Llama 3.2 1B Instruct model, a widely recognized large language model (LLM) released in September 2024.
For evaluation purposes, we utilized the ARC Challenge dataset available from Llama 3.2's official Hugging Face repository.

The ARC dataset for Llama 3.2 comprises 1165 problems presented in natural language.
Each problem prompts the LLM to select the best answer from options A, B, C, and D.
For instance, given a question such as ``What is the main source of energy for all of the organisms in most food chains?'' with possible answers ``A. sunlight, B. water, C. green plants, D. decomposers'', the LLM is expected to identify the correct answer, which is A.

\subsection{Preservation of Functionality}
As a first step, we evaluated the floating-point model processed by \splitquantvii to verify its preservation of functionality.
The evaluation confirmed that the outputs generated by the \splitquantvii-processed floating-point model were identical to those produced by the original model across all 1165 problems in the ARC Challenge dataset, as anticipated.

\subsection{\splitquantvii improves the accuracy of the quantization model}
\begin{table}
\centering
\begin{tabular}{crrr}
\toprule
Llama 3.2 & Baseline & \splitquantvii & Diff.\\
1B Instruct & & &\\
\midrule
Original & 57.94\% & 57.94\% & 0.0\%p\\
INT8     & 57.85\% & 57.85\% & 0.0\%p\\
INT4     & 45.92\% & 57.68\% & \underline{\textbf{+11.76\%p}}\\
INT2     &  0.0\%  &  0.0\%  & 0.0\%p\\
\bottomrule
\end{tabular}
\caption{INT8 linear quantization is not much affected by outliers due to its high quantization resolution. In contrast, INT4 linear quantization reduced accuracy from 57.94\% to 45.92\%. Applying \splitquantvii to INT4 linear quantization improved accuracy by 11.76\%p, nearly matching the original model's accuracy. At INT2, both the baseline and \splitquantvii-enhanced models showed zero accuracy due to extremely low quantization resolution.}
\label{tab:eval}
\end{table}

To evaluate the impact of \splitquantvii, the Llama 3.2 1B Instruct model was linearly quantized to INT2, INT4, and INT8, both with and without the application of \splitquantvii.
The results are presented in Table~\ref{tab:eval}

INT8 linear quantization demonstrated strong performance due to its relatively high quantization resolution, which reduces the influence of outliers.

In contrast, INT4 linear quantization resulted in a decrease in accuracy from the original 57.94\% to 45.92\%.
However, applying \splitquantvii to INT4 significantly improved accuracy by 11.76\%p, nearly matching the accuracy of the original floating-point model.

At INT2, the quantization resolution was so reduced that both the baseline and \splitquantvii-enhanced models exhibited zero accuracy.
This behavior aligns with observations reported by GPTQ, which noted that basic rounding schemes perform adequately up to INT4 quantization but experience severe degradation from INT3 onward.
Additionally, INT2 quantized models generated output text strings consisting of random characters.

\subsection{Running Time of \splitquantvii}
The running time of \splitquantvii was evaluated on an Apple M4 CPU in CPU-only mode.
On average, applying \splitquantvii to the Llama 3.2 1B Instruct model took 1 minute and 58 seconds.
An additional 8 seconds were required to linearly quantize the model processed by \splitquantvii, resulting in a total quantization time of 2 minutes and 6 seconds.

It is noteworthy that this speed was achieved using only the CPU.
Compared to other LLM-specific quantization algorithms that utilize GPUs, this is remarkably fast.
For instance, ZeroQuant requires 3.1 hours to quantize a 1.3B LLM on an A100 GPU~\cite{yao2022zeroquant}.
Although GPTQ is considerably faster, taking 2.9 minutes to quantize a 1.7B LLM, it too was measured using an A100 GPU~\cite{frantar2022gptq}.

\section{Limitation and Future Work}
One of the primary limitations of \splitquantvii is that it increases the size, memory usage and inference time of the quantized model.
For example, while quantizing an FP32 model to INT4 typically reduces the model size to 1/8 of its original, the application of \splitquantvii for INT4 quantization results in a model size of 3/8 of the original.
Although this is a reduction compared to the original model, achieving further size reduction would be advantageous.
Additionally, the increased number of layers leads to longer inference times.

To address this, one potential strategy is to reduce the number of clusters from three to two.
While this may decrease the quantization resolution compared to using three clusters, it offers a trade-off between improved accuracy and reduced model size.
A more sophisticated approach involves dynamically determining the number of clusters for each layer, allowing for flexibility based on the distribution of values within those layers.
However, this method may involve additional computations, potentially increasing the running time.

Furthermore, \splitquantvii currently supports only weight quantization, as it is designed for situations where a calibration dataset is unavailable.
In contrast, when a calibration dataset is accessible, it can be used to simulate the output values of the activation layer.
Then by employing k-means clustering on these simulated activation values, the activation layer can be effectively partitioned.
Employing masking layers to selectively activate or deactivate values based on their respective clusters will be useful for splitting the activation layer.
This technique could enhance the quantization resolution of activation layers, thereby further improving the performance of the quantized model.

\section{Conclusion}
\splitquantvii offers a novel approach to restructuring LLMs to achieve effective quantization using basic linear quantization, even in the absence of GPUs or calibration datasets.
The method involves applying k-means clustering to the weights of layers.
This clustering process enables the original layer to be split into three separate layers while maintaining its original functionality, thereby enhancing the quantization resolution.

Our evaluation was conducted using the Llama 3.2 1B Instruct model and the ARC Challenge dataset.
It was demonstrated that \splitquantvii improves the accuracy of INT4 linear quantization by 11.72\%p, closely approximating the accuracy of the original floating-point model.
Notably, \splitquantvii is both fast and simple to implement; it can be applied to the Llama 3.2 1B Instruct model in approximately 2 minutes using an Apple M4 CPU.
This efficiency makes it particularly useful in environments lacking powerful GPUs.

To the best of our knowledge, \splitquantvii is the first technique to restructure LLMs specifically to enhance quantization results without the need for GPUs.
We believe that \splitquantvii will significantly increase the accessibility of LLMs and their quantization in GPU-limited settings.
The tool is available for download at its online repository.\footnote{The URL of the online repository is currently hidden for blind review because it contains the name of the author(s).}

\bibliographystyle{named}
\bibliography{splitquantv2}

\end{document}